\documentclass[lettersize,journal]{IEEEtran}
\usepackage{amsmath,amsfonts}
\usepackage{algorithmic}
\usepackage{algorithm}
\usepackage{array}
\usepackage[caption=false,font=normalsize,labelfont=sf,textfont=sf]{subfig}
\usepackage{textcomp}
\usepackage{stfloats}
\usepackage{url}
\usepackage{verbatim}
\usepackage{graphicx}
\usepackage{cite}
\usepackage{multirow}
\usepackage{hyperref}
\usepackage{arydshln}
\hypersetup{hypertex=true,
            colorlinks=true,
            linkcolor=blue,
            anchorcolor=blue,
            citecolor=blue}
\begin{document}

\title{CrossTracker: Robust Multi-modal 3D Multi-Object Tracking via Cross Correction}

\author{Lipeng Gu, Xuefeng Yan, Weiming Wang, Honghua Chen, Dingkun Zhu, Liangliang Nan, and Mingqiang Wei, \textit{Senior Member}, \textit{IEEE}
\thanks{L. Gu, H. Chen and M. Wei are with School
of Computer Science and Technology, Nanjing University of Aeronautics
and Astronautics, Nanjing, China (e-mail: 
glp1224@163.com; chenhonghuacn@gmail.com; mingqiang.wei@gmail.com).}
\thanks{X. Yan is with the School of Computer Science and Technology, Nanjing University of Aeronautics and Astronautics, Nanjing, China, and also with the Collaborative Innovation Center of Novel Software Technology and Industrialization, Nanjing, China (e-mail: yxf@nuaa.edu.cn).}
\thanks{W. Wang is with the School of Science and Technology, Hong Kong Metropolitan University, Hong Kong SAR (e-mail: wmwang@hkmu.edu.hk).}
\thanks{D. Zhu is with the School of Computer Science, Jiangsu University of Technology, Changzhou, China (e-mail: zhudingkun@jsut.edu.cn).}
\thanks{L. Nan is with the Urban Data Science Section, Delft University of Technology, Delft, Netherlands (e-mail: liangliang.nan@tudelft.nl).}
}
\markboth{Journal of \LaTeX\ Class Files,~Vol.~14, No.~8, August~2021}%
{Shell \MakeLowercase{\textit{et al.}}: A Sample Article Using IEEEtran.cls for IEEE Journals}


\maketitle

\begin{abstract}
The fusion of camera- and LiDAR-based detections offers a promising solution to mitigate tracking failures in 3D multi-object tracking (MOT).
However, existing methods predominantly exploit camera detections to correct tracking failures caused by potential LiDAR detection problems, neglecting the reciprocal benefit of refining camera detections using LiDAR data.
This limitation is rooted in their single-stage architecture, akin to single-stage object detectors, lacking a dedicated trajectory refinement module to fully exploit the complementary multi-modal information.
To this end, we introduce \textit{CrossTracker}, a novel two-stage paradigm for online multi-modal 3D MOT.
CrossTracker operates in a coarse-to-fine manner, initially generating coarse trajectories and subsequently refining them through an independent refinement process.
Specifically, CrossTracker incorporates three essential modules: 
i) a multi-modal modeling (M$^3$) module that, by fusing multi-modal information (images, point clouds, and even plane geometry extracted from images), provides a robust metric for subsequent trajectory generation.
ii) a coarse trajectory generation (C-TG) module that generates initial coarse dual-stream trajectories, 
and iii) a trajectory refinement (TR) module that refines coarse trajectories through cross correction between camera and LiDAR streams.
Comprehensive experiments demonstrate the superior performance of our CrossTracker over its eighteen competitors, underscoring its effectiveness in harnessing the synergistic benefits of camera and LiDAR sensors for robust multi-modal 3D MOT.
\end{abstract}


\section{Introduction}

3D multi-object tracking (MOT), essential for accurately interpreting object motion trajectories in 3D space, has become an indispensable component of various robotic applications \cite{VILENS,wensing2023optimization,xie2021unseen}, including autonomous vehicles, indoor robots, and industrial robots.
Existing methods can be categorized into single-modal and multi-modal paradigms.
Single-modal methods \cite{ab3dmot,centertrack,centerpoint,streampetr} typically employ camera- or LiDAR-based 3D detectors to detect objects frame by frame, followed by tracking them over time. However, object detection from a single-modal source is typically not reliable and inevitably leads to missing or false detections, which directly hinder the subsequent 3D MOT performance.

\begin{figure}[t]
 \centering
  \includegraphics[width=0.48\textwidth]{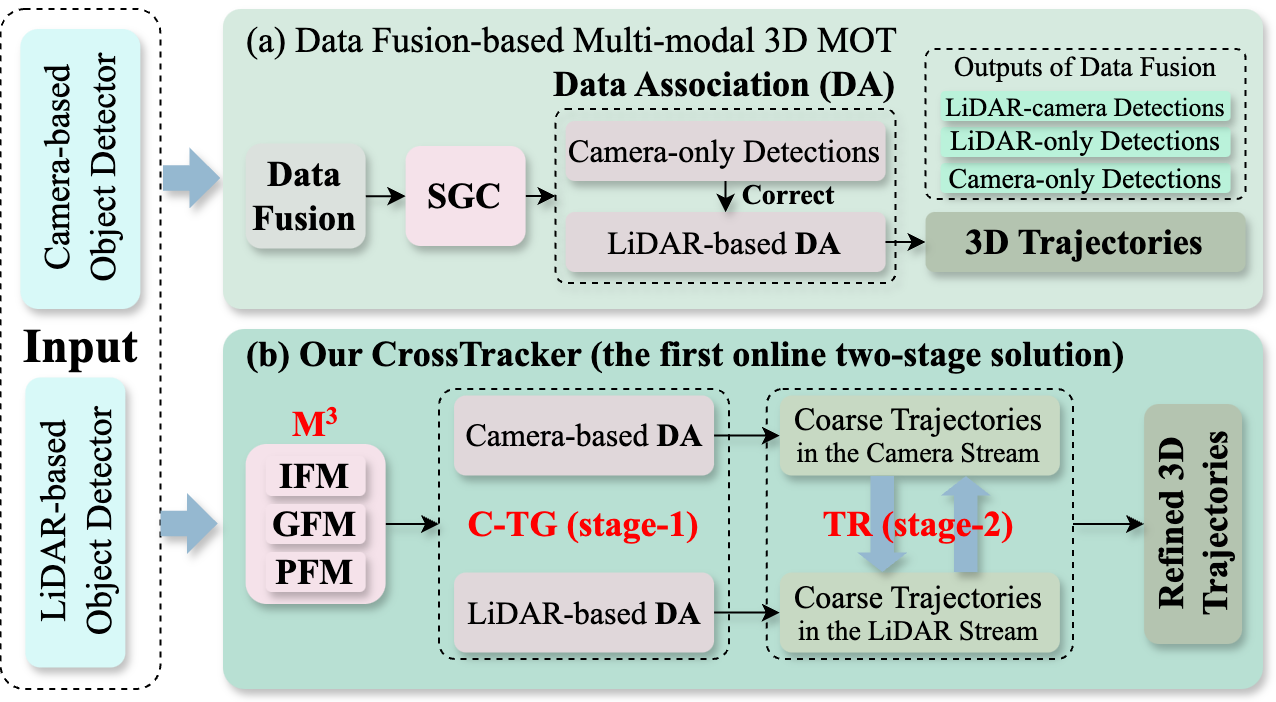}
  \caption{\textbf{Main difference between CrossTracker and its competitors.}
  Prior methods \cite{eagermot,deepfusionmot,sfmot} typically categorize input detections into three sets using data fusion, and then sequentially process these detection sets based on spatial geometric constraints (SGC, e.g., 3D-IoU). They typically start with LiDAR-camera detections, followed by LiDAR-only detections. Finally, they use camera-only detections to correct tracking failures caused by potential LiDAR detection problems.
  However, they are unable to correct tracking failures in the camera stream using their single-stage architecture.
  Differently, our CrossTracker, the first online two-stage 3D MOT solution, excels in addressing intricate tracking failures in both streams. 
  It leverages multi-modal modeling (M$^3$), encompassing image, geometric, and point cloud feature modeling (IFM, GFM, and PFM), followed by a two-stage tracking pipeline consisting of coarse trajectory generation (C-TG) and trajectory refinement (TR).
  }
  \label{fig:pipeline}
\end{figure}

\begin{figure*}[ht]
 \centering
  \includegraphics[width=1\textwidth]{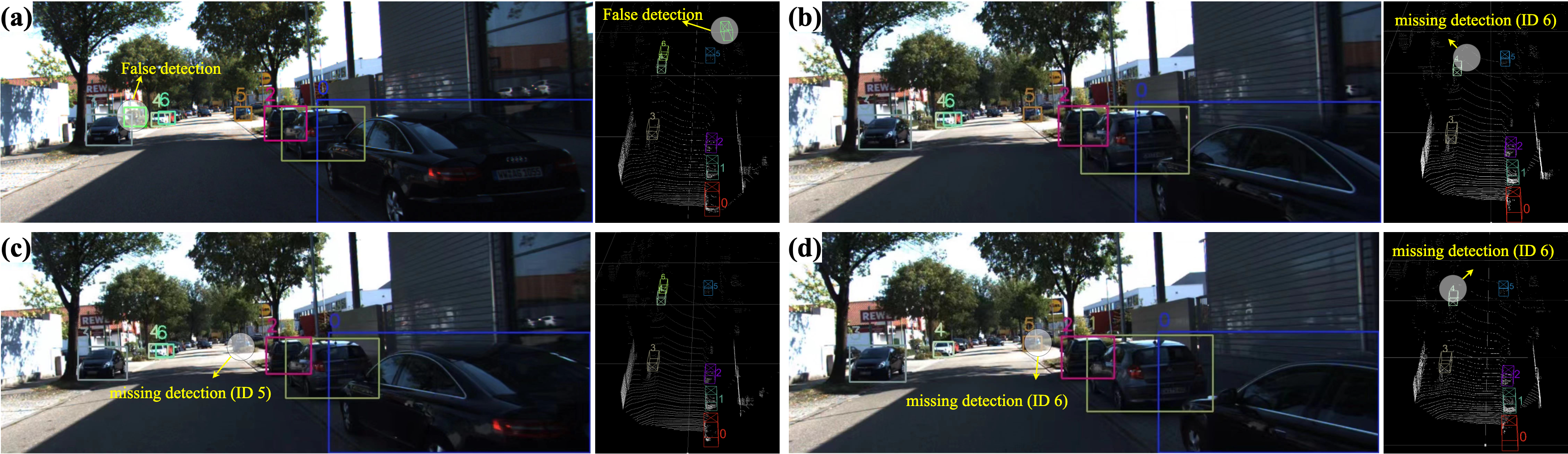}
  \caption{\textbf{The advantages and disadvantages of existing data fusion-based multi-modal 3D MOT methods \cite{eagermot,deepfusionmot,sfmot}.}
  These methods can effectively address (a) false detections in either the camera or LiDAR stream and (b) missing detections solely in the LiDAR stream. However, they are limited in their ability to handle (c) missing detections solely in the camera stream and (d) missing detections in both camera and LiDAR streams.
  In contrast, our CrossTracker draws on the coarse-to-fine concept, effectively addressing all four of these challenges through an innovative two-stage tracking scheme.}
  \label{fig:problems}
\end{figure*}

To mitigate the challenges mentioned above, as shown in Fig. \ref{fig:pipeline} (a), several data fusion-based multi-modal methods \cite{eagermot,deepfusionmot,sfmot} are proposed.
These methods exploit the camera stream as an auxiliary source to correct tracking failures arising from missing and false LiDAR detections.
Initially, reliable LiDAR-camera detections (from both camera and LiDAR streams) are associated with historical trajectories. Subsequently, LiDAR-only detections are matched with the remaining unassociated historical trajectories.
This two-step process minimizes the introduction of false LiDAR detections and reduces the noise interference of false camera detections on LiDAR trajectories (see Fig. \ref{fig:problems} (a)).
Finally, to ensure the continuity of the LiDAR trajectory, camera-only detections are incorporated to bridge the gaps caused by missed LiDAR detections, as illustrated in Fig. \ref{fig:problems} (b).
Despite these advancements, this unidirectional camera-to-LiDAR correction approach faces limitations in challenging scenarios—particularly when there are missing detections in the camera stream (Fig. \ref{fig:problems} (c)) or in both streams (Fig. \ref{fig:problems} (d)).
A natural inquiry arises: why is it essential to correct detection problems in the auxiliary camera stream?
The rationale is straightforward. Corrected camera data from the current frame can more effectively refine LiDAR data in the subsequent frame. 
This iterative process can significantly enhance the robustness of the output LiDAR trajectories.

\textit{What are the root causes of the limitations in existing methods \cite{eagermot,deepfusionmot,sfmot}? Is it the lack of consideration for detection problems in the camera stream, or is it an inherent flaw in the tracking architecture itself?}
Upon closer investigation, we find that these methods merely process different detection sets sequentially according to their design, lacking a dedicated trajectory refinement module to fully exploit the complementary information between camera and LiDAR data for cross correction.
Consequently, similar to the distinction between single-stage detectors (e.g., SSD \cite{ssd}) and two-stage detectors (e.g., Faster R-CNN \cite{fasterrcnn}) in object detection based on the presence of detection refinement modules, these methods can also be categorized as single-stage architectures.
Moreover, their reliance on spatial geometric constraints (e.g., 3D-IoU) introduces hyperparameters that require empirical tuning. This can be cumbersome and less user-friendly, especially when dealing with diverse object categories and datasets.

We propose \textbf{CrossTracker}, an online multi-modal 3D MOT paradigm comprising three essential modules: a multi-modal modeling (M$^3$) module, a coarse trajectory generation (C-TG) module, and a trajectory refinement (TR) module.

To our knowledge, CrossTracker is the first two-stage solution for online 3D MOT: \textit{coarse dual-stream trajectory generation} followed by \textit{trajectory refinement}.
This innovative design, incorporating an additional trajectory refinement phase, distinguishes itself from existing data fusion-based methods \cite{eagermot,deepfusionmot,sfmot}.
Through cross correction between camera and LiDAR streams, CrossTracker is capable of addressing false and missing detections in both modalities, resulting in more robust LiDAR trajectories.
Moreover, the M$^3$ module models discriminative multi-modal features, including joint image features, point cloud features, and even plane geometric features extracted from images, to accurately estimate the consistency probability between object pairs, thereby providing a robust metric for subsequent trajectory generation.

We evaluate the performance of our CrossTracker by conducting a comparative analysis with eighteen competitors on the KITTI tracking benchmark. 
Notably, CrossTracker surpasses all the competitors, demonstrating its superior performance. The contributions are three-fold:
\begin{itemize}
\item We propose \textit{CrossTracker}, the first two-stage 3D MOT solution that significantly improves tracking robustness. It adopts a coarse-to-fine tracking scheme to correct tracking failures caused by various detection problems.
\item We introduce M$^3$, a multi-modal modeling module that leverages both camera and LiDAR data to estimate the consistency probabilities between object pairs, providing a robust metric for subsequent trajectory generation.
\item We propose a two-stage tracking scheme: i) C-TG generates initial coarse dual-stream trajectories; ii) TR refines these trajectories through cross correction, leveraging the complementary strengths of camera and LiDAR data.
\end{itemize}

\section{Related Works}
\subsection{Tracking-by-Detection}
Existing MOT methods can be classified into two main paradigms: tracking-by-detection (TBD) \cite{sort,deepsort,ab3dmot,eagermot,deepfusionmot,sfmot,fantrack} and joint-detection-tracking (JDT) \cite{strongsort,fantrack,fairmot,centerpoint,centertrack} paradigms.
TBD methods perform frame-by-frame detection using off-the-shelf object detectors and subsequently track the resulting detections. In contrast, JDT methods jointly optimize detection and tracking tasks within an end-to-end network.

The TBD paradigm has gained widespread adoption due to its ability to directly leverage cutting-edge object detectors, significantly improving tracking performance.
Sort \cite{sort} is the pioneering work in this paradigm. It adopts a Kalman filter for object state estimation and employs the Hungarian algorithm with 2D-IOU as a data association metric to establish correspondences between trajectories and detections.
DeepSORT \cite{deepsort} extends SORT by incorporating image features, modeled using deep neural networks, to enhance object discrimination.
Subsequent TBD-based MOT methods \cite{fairmot,jrmot,jmodt} have largely built upon the foundation laid by DeepSORT.

With the flexibility of TBD as a foundation, the innovative two-stage architecture of our Crosstracker delivers significantly improved tracking performance.

\begin{figure*}[t]
 \centering
  \includegraphics[width=\textwidth]{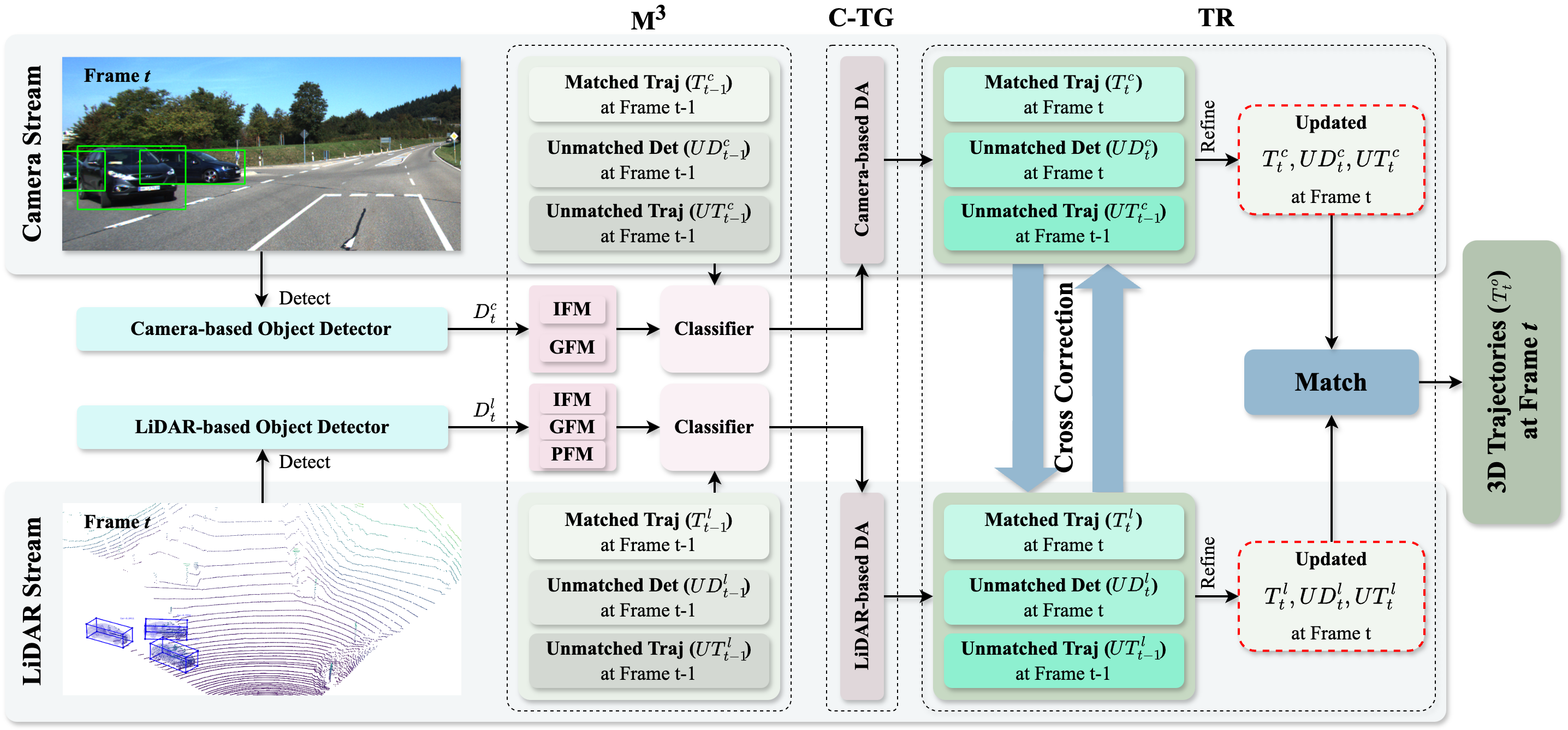}
  \caption{\textbf{Overview of CrossTracker.}
  It is the first two-stage multi-modal 3D MOT framework comprising three essential modules: a multi-modal modeling (M$^3$) module, a coarse trajectory generation (C-TG, i.e., stage-1) module, and a trajectory refinement (TR, i.e., stage-2) module.
  At each frame after the initial frame (e.g., frame $t-1$), trajectories within the camera or LiDAR stream can be categorized into matched trajectories ($T^{k}_{t-1}$), unmatched detections ($UD^{k}_{t-1}$), and unmatched trajectories ($UT^{k}_{t-1}$), where $k\in \left \{ c, l \right \}$ denotes the camera or LiDAR stream, respectively.
  Based on this, CrossTracker streamlines the two-stage 3D MOT problem by addressing these sets from both streams.
  Specifically, M$^3$ is designed with an image feature model (IFM), a point cloud feature model (PFM), and a plane geometric feature model (GFM), coupled with a classifier, to establish a robust metric for subsequent trajectory generation.
  Given the input camera detections ($D^{c}_{t}$) at frame $t$, C-TG initially associates$D^{c}_{t}$ with these sets ($T^{c}_{t-1}$, $UD^{c}_{t-1}$, and $UT^{c}_{t-1}$) at frame $t-1$, respectively.
  The same procedure is applied to the input LiDAR detections ($D^{l}_{t}$) at frame $t$.
  Following this, both streams yield updated matched trajectories ($T^{c}_{t}$ and $T^{l}_{t}$), remaining unmatched detections ($UD^{c}_{t}$ and $UD^{l}_{t}$), and remaining unmatch trajectories ($UT^{c}_{t-1}$ and $UT^{l}_{t-1}$) at frame $t$.
  Subsequently, TR implements cross correction to address tracking failures arising from potential false detection problems in $UD^{c}_{t}$ and $UD^{l}_{t}$ as well as missed detection problems in $UT^{l}_{t-1}$ and $UT^{c}_{t-1}$.
  Finally, high-quality 3D trajectories ($T^{o}_{t}$) is outputted by matching the updated $T^{l}_{t}$ with the updated $T^{c}_{t}$, $UD^{c}_{t}$, and $UT^{c}_{t}$.
  }
  \label{fig:overview}
\end{figure*}


\subsection{3D Multi-object Tracking}
With the development of 3D perception technology, numerous attempts \cite{TuSimple,ab3dmot,centerpoint,eagermot,deepfusionmot,sfmot,mmmot,muller2021seeing,CFTrack} have been made in 3D MOT based on LiDAR or stereo. 
AB3DMOT \cite{ab3dmot} extends SORT \cite{sort} to the 3D domain by employing 3D Kalman filtering to construct a simple yet highly efficient 3D motion model.
Beyond the TBD paradigm, methods based on the JDT paradigm, such as CenterPoint \cite{centerpoint} and JRMODT \cite{jmodt}, have also demonstrated promising results. CenterPoint exclusively utilizes the point cloud for object detection and employs the predicted center velocity of each object for data association. Despite variations in paradigms, these methods are limited by the use of a single data source. The input detections are prone to both missed and false detections, resulting in unstable trajectories.
Recent data fusion-based multi-modal methods \cite{eagermot,deepfusionmot,sfmot} capitalize on the complementary strengths of LiDAR and camera sensors.
However, they only consider using camera information to correct LiDAR trajectories, overlooking the fact that corrected camera data from the current frame can more effectively refine LiDAR data in the subsequent frame. This limitation stems from their single-stage architecture, which restricts their ability to fully exploit complementary multi-modal information for cross correction.

Our CrossTracker, a novel two-stage tracking framework, combines coarse trajectory generation and refinement stages. By leveraging cross correction between camera and LiDAR streams, our method produces more robust 3D trajectories.

\begin{figure*}[t]
 \centering
  \includegraphics[width=\textwidth]{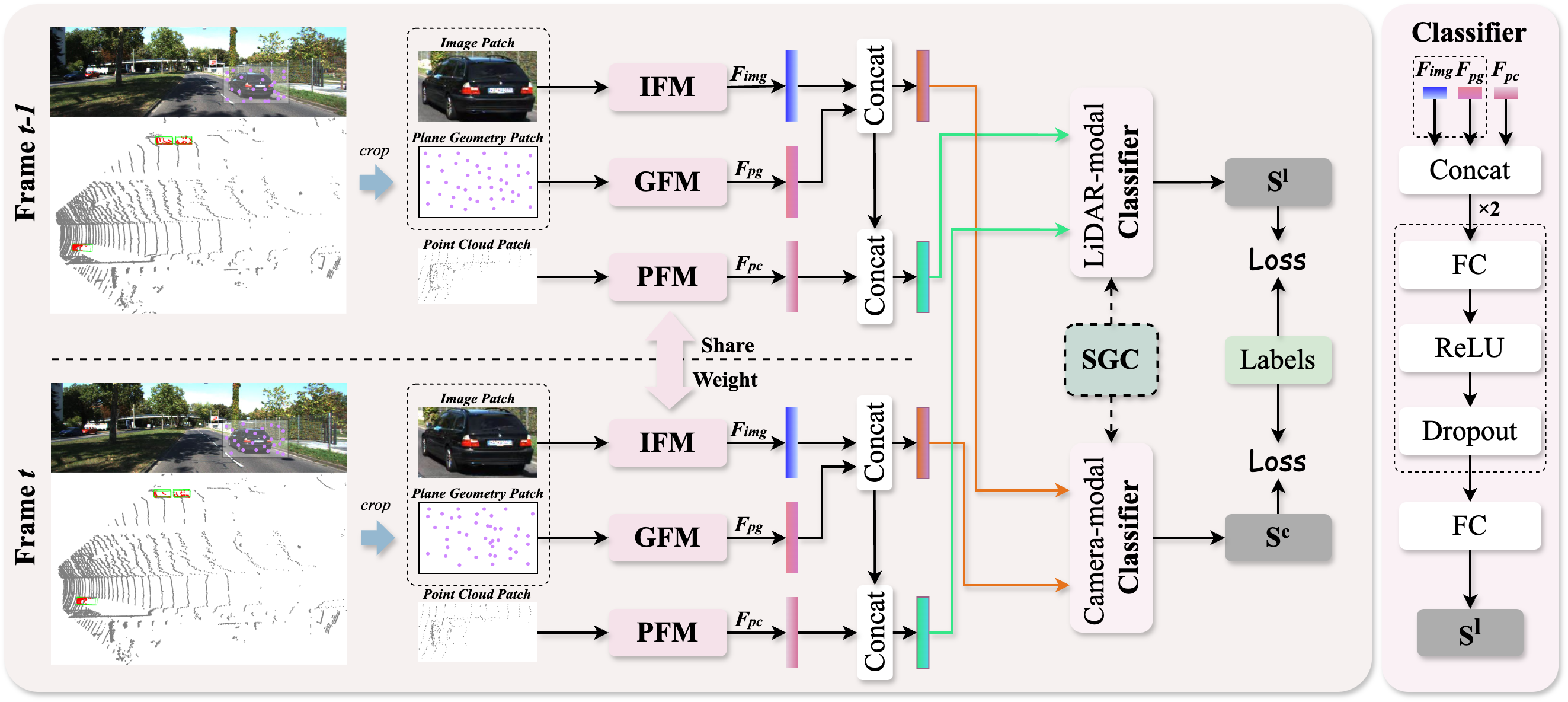}
  \caption{\textbf{Overview of M$^3$ module.} 
  It takes as input two consecutive frames of the image and point cloud, along with their corresponding detections. It independently outputs the consistency probabilities (similarity scores) of two objects for the camera ($S^{c}$) and LiDAR ($S^{l}$) scenario.
  M$^3$ comprises four primary components: 
  the image feature modeling (IFM) module for image features ($F_{img}$), 
  the plane geometric feature modeling (GFM) module for plane geometry features ($F_{pg}$) and 
  the point cloud feature modeling (PFM) module for point cloud features ($F_{pc}$), and the camera- and LiDAR-modal classifiers for estimating consistency probabilities.
  Furthermore, SGC incorporates spatial geometric constraints, such as 3D centroid distance, during inference to further refine the consistency probability output by classifiers.
  }
  \label{fig:M^3}
\end{figure*}

\section{Method}

\subsection{Overview}
CrossTracker is the first online two-stage multi-modal 3D MOT solution that leverages multi-modal information from both the current and historical frames.
As shown in Fig. \ref{fig:overview}, it comprises three modules: (1) a multi-modal model (M$^3$) module; (2) a coarse trajectory generation (C-TG, i.e., stage-1) module; (3) a trajectory refinement (TR, i.e., stage-2) module. 

At each frame after the initial frame (e.g., frame $t-1$), trajectories within the camera ($k=c$) or LiDAR ($k=l$) stream can be categorized into matched trajectories ($T^{k}_{t-1}$), unmatched detections ($UD^{k}_{t-1}$), and unmatched trajectories ($UT^{k}_{t-1}$).
Building upon this, CrossTracker simplifies 3D MOT by employing a two-stage, coarse-to-fine approach to refine these sets from both streams. 
In the first stage (i.e., intra-modal tracking), input detections ($D^{k}_{t}$) are associated with these three sets ($T^{k}_{t-1}$, $UD^{k}_{t-1}$, and $UT^{k}_{t-1}$) using the robust metric provided from the M$^3$ module.
In the second stage (i.e., cross-modal correction), cross correction is implemented to address tracking failures arising from potential false detection problems in $UD^{l}_{t}$ and $UD^{c}_{t}$ as well as missed detection problems in $UT^{l}_{t-1}$ and $UT^{c}_{t-1}$ at frame $t$.
Finally, high-quality 3D trajectories ($T^{o}_{t}$) is outputted by matching the updated $T^{l}_{t}$ with the updated $T^{c}_{t}$, $UD^{c}_{t}$, and $UT^{c}_{t}$.

\subsection{Multi-modal Modeling}
To achieve more robust performance, existing methods \cite{deepsort,eagermot,deepfusionmot,jmodt,mmmot,fantrack} typically focus on modeling more discriminative features for each detected object.
Single-modal methods \cite{sort,ab3dmot,deepfusionmot,eagermot,centerpoint,centertrack} often employ carefully crafted spatial geometric constraints (SGCs, e.g., 3D-IoU) for tracking.
However, they are susceptible to tracking failures when dealing with objects that have similar positions but distinct appearances.
To address this limitation, multi-modal methods \cite{mmmot,deepsort,fairmot} incorporate deep neural networks (DNNs) to extract more comprehensive image and point cloud features, complementing SGCs.
However, they often overlook the integration of geometric information into DNNs for end-to-end fusion with image and point cloud features, which could simplify multi-modal feature modeling.

To address the aforementioned limitations, we propose M$^3$, a multi-modal modeling network. 
As illustrated in Fig. \ref{fig:M^3}, M$^3$ is the first to integrate image, plane geometry, and point cloud modeling, along with consistency probabilistic estimation, into a unified end-to-end network.
Specifically, M$^3$ includes four components: an image feature modeling (IFM) module for image features ($F_{img}$), a plane geometric feature modeling (GFM) module for plane geometry features ($F_{pg}$), a point cloud feature modeling (PFM) for point cloud features ($F_{pc}$), a classifier module for estimating consistency probability, and a spatial geometric constraints (SGC) module for further refining this estimation during inference.
Notably, the M$^3$ module can also be configured to exclusively use the camera sensor for modeling image features and plane geometric features.

\subsubsection{Image Feature Modeling (IFM)}
The IFM module is instrumental in image feature modeling. To ensure real-time performance, we employ a \textit{ResNet-18} \cite{resnet} with a \textit{MaxPooling} layer. The module takes an image patch as input and outputs a compact feature vector.
Specifically, an input image patch is cropped from a specific region of the image based on the provided 2D bounding box, which may originate from a 2D detection or the projection of a 3D detection onto the image plane. The patch is then resized to $80 \times 80$ pixels.
The IFM module processes the image patch to extract an image feature vector, denoted as $F_{img} \in \mathbb R ^ {1 \times 512}$.

\subsubsection{Point Cloud Feature Modeling (PFM)}
The PFM module is a cornerstone for point cloud feature modeling. Leveraging the lightweight and efficient \textit{PointNet} \cite{pointnet} with a \textit{MaxPooling} layer, PFM effectively captures and represents the intricate characteristics of point clouds.
Specifically, PFM takes as input a point cloud patch, denoted $P \in \mathbb R ^{N \times c}$, where $N$ and $c$ are the number of points and channels, respectively.
This patch is obtained using 3D bounding boxes and then re-sampled to a fixed size of 512 points. The module subsequently outputs a point cloud feature vector, denoted as $F_{pc} \in \mathbb R ^ {1 \times 512}$.

\subsubsection{Plane Geometric Feature Modeling (GFM)}
In terms of the geometry information, FANTrack \cite{fantrack} pioneers the extraction of geometric information by directly feeding bounding box parameters into convolutional neural networks.
However, relying solely on center coordinates and side lengths limits feature expressiveness and interpretability.
To address this, we propose a novel approach that transforms bounding boxes into intuitively understandable pseudo-point patches, denoted $P \in \mathbb R ^{N \times 2}$.
These patches are generated by randomly sampling 512 points within the 2D bounding box, which can represent either a 2D detection or the projection of a 3D detection onto the image plane. Each point is encoded as a two-dimensional vector, with the two dimensions representing the normalized horizontal and vertical coordinates on the image plane.
The patch is directly fed into a \textit{PointNet} with a \textit{MaxPooling} layer, outputting a feature vector, denoted as $F_{pg} \in \mathbb R ^ {1 \times 512}$.

\begin{figure*}[t]
 \centering
  \includegraphics[width=\textwidth]{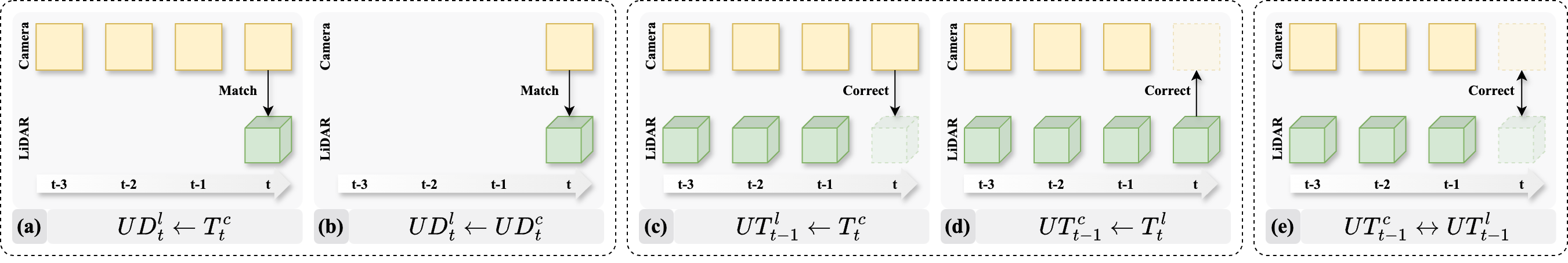}
  \caption{
  \textbf{Five cases of the cross correction in TR.} 
  Each camera and LiDAR stream exhibits a trajectory. Solid shapes (squares for camera, cubes for LiDAR) represent detected objects, while dashed shapes indicate missed detections.
  (a) and (b) identify newly appearing objects from unmatched LiDAR detections in $UD^{l}_{t}$ using camera data as a reference, with a focus on minimizing false detections.
  (c) and (d) correct unmatched camera or LiDAR trajectories in $UT^{c}_{t-1}$ or $UT^{l}_{t-1}$ using the other modality.
  (e) corrects unmatched trajectories in $UT^{c}_{t-1}$ and $UT^{l}_{t-1}$ that may result from simultaneous misses in both modalities.
  }
  \label{fig:examples of cr}
\end{figure*}

\subsubsection{Classifier}
We recast the complex task of estimating object consistency probability as a binary classification problem: determining whether two objects from different frames correspond to the same object, using \textit{Cross Entropy Loss}.
As depicted in Fig.~\ref{fig:M^3}, adhering to a lightweight design, we construct the \textit{Classifier} using only three \textit{FC} layers, along with two \textit{Dropout} and two \textit{ReLU} layers.
Specifically, the classifier takes as input three feature vectors, denoted as $concat(F_{img}, F_{pg}, F_{sg}) \in \mathbb R^{1 \times 1536} $, and outputs the consistency probability $S^{l}$.
Additionally, our classifier also supports the camera scenario. In this case, it takes as input two feature vectors, denoted as $concat(F_{img}, F_{pg}) \in \mathbb R^{1 \times 1024} $, and outputs the consistency probability $S^{c}$.
Notably, the classifier shares the same architecture across LiDAR and camera scenarios, yet weights are not shared.
Given the output consistency probability from the classifier, the cost matrix of multi-modal features between the $M_{t-1}$ objects (including $T_{t}$, $UD_{t-1}$ and $UT_{t-1}$) in the previous frame $t-1$ and the $N_{t}$ detected objects ($D_{t}$) in the current frame $t$ is calculated as
\begin{equation}
S^{k}=
\begin{bmatrix}
S^{k}_{1,1} & S^{k}_{1,2}  & \cdots   & S^{k}_{1,N^{k}_{t}}   \\
S^{k}_{2,1} & S^{k}_{2,2}  & \cdots   & S^{k}_{2,N^{k}_{t}}  \\
\vdots & \vdots  & \ddots   & \vdots  \\
S^{k}_{M^{k}_{t-1},1} & S^{k}_{M^{k}_{t-1},2}  & \cdots\  & S^{k}_{M^{k}_{t-1},N^{k}_{t}}  \\
\end{bmatrix}  \label{cost matrix1}
\end{equation}
where $k\in \left \{ c, l \right \}$ denotes the camera or LiDAR stream, respectively. $i \in \left\{1, ..., M_{t-1}\right\}$, and $j \in \left\{1, ..., N_{t}\right\}$.

\subsubsection{Spatial Geometric Constraint (SGC)}
To further enhance the discriminativeness of features from the $M^{3}$ module, we incorporate the spatial geometric constraint (SGC) during inference to refine the cost matrix $S$ (see Eq. \ref{cost matrix1}). 
\textit{Empirically}, we leverage the 2D Intersection over Union (2D-IoU) and the 3D centroid distance (3D-CD) for 2D detections and 3D detections, respectively, to measure the positional distance between objects on the image plane or 3D spaces. 
To compensate for inter-frame displacements, we adopt Kalman Filtering (KF) to predict historical 2D/3D trajectories, a widely popular method due to its simplicity and effectiveness in various tracking applications.
The KF used here is identical to that in SORT \cite{sort} and AB3DMOT \cite{ab3dmot}, and will not be elaborated on further.
Similar to Eq. \ref{cost matrix1}, the cost matrix of SGC between adjacent frames is defined as
\begin{equation}
\begin{split}
G^{k}=
\begin{bmatrix}
G^{k}_{1,1} & G^{k}_{1,2}  & \cdots   & G^{k}_{1,N^{k}_{t}}   \\
G^{k}_{2,1} & G^{k}_{2,2}  & \cdots   & G^{k}_{2,N^{k}_{t}}  \\
\vdots & \vdots  & \ddots   & \vdots  \\
G^{k}_{M^{k}_{t-1},1} & G^{k}_{M^{k}_{t-1},2}  & \cdots\  & G^{k}_{M^{k}_{t-1},N^{k}_{t}}  \\
\end{bmatrix} \label{cost matrix2}
\end{split}
\end{equation}
where $k\in \left \{ c, l \right \}$. $G^{c}$ is the 2D-IoU metric for 2D detections, and $G^{l}$ is the 3D-CD metric for 3D detections. If $G^{k}$ is the 2D-IoU metric, then set $G^{k}$ to 1 minus $G^{k}$.

\subsection{Stage-1: Coarse Trajectory Generation (C-TG)}
Unlike prior methods \cite{eagermot,deepfusionmot,sfmot} which typically adopt a single-stage tracking framework (see Fig.~\ref{fig:pipeline} (a)), our CrossTracker marks the pioneering adoption of a two-stage framework (see Fig. \ref{fig:overview}).
Analogous to two-stage object detectors \cite{fasterrcnn,pointrcnn} which employs a region proposal network to generate coarse candidate boxes for subsequent refinement, CrossTracker proposes a coarse trajectory generation (C-TG) module to generate coarse trajectories for both camera and LiDAR streams. Subsequently, a trajectory refinement (TR) module is designed to iteratively refine these coarse trajectories through cross correction between camera and LiDAR streams.

Within the C-TG module, $N_{t}$ detections in $D_{t}$ at frame $t$ are initially ranked in descending order based on their detection scores. 
Subsequently, a total cost matrix $C^{k}$ is constructed to quantify the association costs between $N_{t}$ detections at frame $t$ and $M_{t-1}$ trajectories in $T_{t-1}$, $UD_{t-1}$ and $UT_{t-1}$ at frame $t-1$, where $k\in \left \{ c, l \right \}$.
Specifically, the association cost between the $i-$th object at frame $t-1$ and the $j-$th object at frame $t$, denoted by $C^{k}_{i,j}$, is defined as
\begin{equation} \label{total cost matrix}
\begin{split}
C^{k}_{i,j}=\left\{
 \begin{array}{ll}
S^{k}_{i,j} + G^{k}_{i,j} &, condition1 \\
 1000 &, condition2 \\
 \end{array}
 \right.
\end{split}
\end{equation}
where 
$condition1$ is $S^{k}_{i,j} \ge \theta_{S} \ or \ G^{k}_{i,j} \le \theta_{G}$, $condition2$ is $S^{k}_{i,j} < \theta_{S} \ and \ G^{k}_{i,j} > \theta_{G}$, and
$\theta_{S}$ and $\theta_{G}$ represent thresholds for the consistency probabilities of multi-modal features and the SGC between object pairs, respectively.
A greedy algorithm is then employed to associate detections with historical trajectories, guided by Eq. \ref{total cost matrix}.
As a result, both camera and LiDAR streams yield three sets of trajectories for each sensor:
coarse matched trajectories $T^{k}_{t}$, unmatched detections $UD^{k}_t$, and unmatched trajectories $UT^{k}_{t-1}$. 
These six sets $T^{c}_{t}$, $T^{l}_{t}$, $UD^{c}_t$, $UD^{l}_t$, $UT^{c}_{t-1}$, and $UT^{l}_{t-1}$, from both the camera and LiDAR streams, are then input to the TR module to refine the dual-stream trajectories through cross correction.

\subsection{Stage-2: Trajectory Refinement (TR)} \label{CR}
Our TR module differs from existing methods such as \cite{eagermot,deepfusionmot,sfmot} in that it can effectively address all the detection problems (see Fig. \ref{fig:problems}) that adversely impact trajectory robustness. This is achieved through the independent trajectory refinement design (i.e., cross correction) of our TR module, which enables seamless interaction between image and point cloud information.
The core of our cross-correction process lies in the identification and transfer of newly appearing objects from $UD^{c}_{t}$ and $UD^{l}_{t}$ into $T^{c}_{t}$ and $T^{l}_{t}$, respectively.
Concurrently, unmatched trajectories from $UT^{c}_{t-1}$ and $UT^{l}_{t-1}$, erroneously omitted due to missed detections, are recovered and transferred into $T^{c}_{t}$ and $T^{l}_{t}$.
Finally, high-quality 3D trajectories ($T^{o}_{t}$) are outputted by matching the corrected LiDAR and camera data.
To systematically address the intricate cross-correction process, we categorize it into five cases (see Fig. \ref{fig:examples of cr}) and process them in three steps.
Prior to cross correction, $T^{l}_{t}$ and  $T^{c}_{t}$ are matched using the 2D-IoU criterion with the threshold $\theta_{iou}$. 
The subsequent cross correction will exclude matched trajectories from both $T^{l}_{t}$ and  $T^{c}_{t}$, but note that these matched trajectories are not removed from their corresponding sets.

\textbf{STEP-1:}
\textit{
Unmatched detections in $UD^{l}_{t}$ represent either false detections or newly appearing objects (see Fig. \ref{fig:examples of cr} (a) and (b)).}
To identify the latter, a greedy algorithm pairs detections in $UD^{l}_{t}$ with those in $T^{c}_{t}$ using 2D-IoU, discarding pairs below threshold $\theta_{iou}$.
Paired detections in $UD^{l}_{t}$ are initialized as new trajectories, added to $T^{l}_{t}$, and then removed from $UD^{l}_{t}$.
Similarly, remaining detections in $UD^{l}_{t}$ and $UD^{c}_{t}$ are paired, initialized as new trajectories, added to $T^{l}_{t}$ and $T^{c}_{t}$, respectively, and then removed from their respective sets.
\textit{Note that any detections in $UD^{l}_{t}$ and $UD^{c}_{t}$ that remain unobserved and uncorrected for $N$ consecutive frames are classified as false detections and discarded.}

\textbf{STEP-2:}
\textit{Unmatched trajectories in $UT^{c}_{t-1}$ and $UT^{l}_{t-1}$ can arise from trajectory termination or missed detections in one or both modalities.}
Given the higher amenability of single-modality missed detections to correction, cases like those in Fig. \ref{fig:examples of cr} (c) and (d) are prioritized.
Initially, the KF model predicts the states of trajectories in $UT^{l}_{t-1}$ at frame $t$, yielding $\hat{UT}^{l}_{t}$. 
Unmatched trajectories in $\hat{UT}^{l}_{t}$ are then compared to those in $T^{c}$, following the procedure in \textbf{STEP-1}.
Beyond the 2D-IoU criterion, we further stipulate that both compared trajectories must have at least $\theta_{hits}$ consecutive observations ($hits$).
For each pair $(\hat{\textbf{ut}}^{l}_{t}, \textbf{ut}^{c}_{t})$ meeting these criteria, the corresponding $\textbf{ut}^{l}_{t-1}$ is updated to frame $t$ using $\hat{\textbf{ut}}^{l}_{t}$, added to $T^{l}_{t}$, and removed from $UT^{l}_{t-1}$.
Similarly, for each paired $(\hat{\textbf{ut}}^{c}_{t}, \textbf{t}^{l}_{t})$, the corresponding $\textbf{ut}^{c}_{t-1}$ is updated to frame $t$ using $\hat{\textbf{ut}}^{l}_{t}$, added to $T^{c}_{t}$, and removed from $UT^{c}_{t-1}$.

\textbf{STEP-3:}
\textit{Remaining unmatched trajectories in $UT^{c}_{t-1}$ and $UT^{l}_{t-1}$ may arise from simultaneous termination or missed detections in both modalities (see Fig. \ref{fig:examples of cr} (e)).}
Based on predicted $\hat{UT}^{c}_{t}$ and $\hat{UT}^{l}_{t}$ at frame $t$ from $UT^{c}_{t-1}$ and $UT^{l}_{t-1}$, unmatched trajectories in both sets are compared using the procedure in \textbf{STEP-2}.
Beyond the 2D-IoU and $hits$ criteria, we further stipulate that both predicted trajectories must not be at the image plane boundary.
This constraint effectively determines whether trajectories have terminated.
For each pair $(\hat{\textbf{ut}}^{c}_{t}, \hat{\textbf{ut}}^{l}_{t})$ meeting these criteria, the corresponding $\textbf{ut}^{c}_{t-1}$ and $\textbf{ut}^{l}_{t-1}$ are updated to frame $t$ using $\hat{\textbf{ut}}^{c}_{t}$ and $\hat{\textbf{ut}}^{l}_{t}$, respectively, added to $T^{c}_{t}$ and $T^{l}_{t}$, and removed from $UT^{c}_{t-1}$ and $UT^{l}_{t-1}$.
\textit{Note that any trajectories in $UT^{c}_{t-1}$ and $UT^{l}_{t-1}$ that remain unobserved and uncorrected for $N$ consecutive frames are deemed terminated and discarded.}

After comprehensive cross correction, a significant reduction in false and missed detections is achieved in both streams at frame $t$.
To further enhance tracking performance, the updated $T^{l}_{t}$ is associated not only with the updated $T^{c}_{t}$ but also with the updated $UD^{c}_{t}$ and $UT^{c}_{t}$. This association is based on the 2D-IoU criterion with the threshold $\theta_{iou}$.
The stricter constraints imposed on $UD^{c}_{t}$ and $UT^{c}_{t}$ during the cross correction process may result in some new detections or reliable unmatched trajectories remaining uncorrected.
As illustrated in case (d) of Fig. \ref{fig:examples of cr}, if the number of hits of the LiDAR trajectory falls below the threshold $\theta_{hits}$, the unmatched camera trajectory in $UT^{c}_{t}$ cannot be corrected. Hence, by incorporating $UD^{c}_{t}$ and $UT^{c}_{t}$ into the association process, more potential matches can be effectively captured.
Finally, high-quality 3D trajectories ($T^{o}_{t}$) are generated by selecting trajectories from $T^{l}_{t}$ that have a match in at least one of the other three sets.

\begin{table*}[t]
\caption{
Comparison of 3D MOT methods on the \textit{Car} and \textit{Pedestrian} categories of the KITTI \textit{test} set. 
Methods superscripted with ``C1'', ``C2'', and ``C3'' use camera-based detectors: RRC \cite{rrc} for cars and Track-RCNN \cite{trackrcnn} for pedestrians; Perma \cite{PermaTrack} for both cars and pedestrians; and Perma for both cars and pedestrians, respectively. 
Methods superscripted with ``L1'', ``L2'', and ``L3'' use LiDAR-based detectors: PointGNN \cite{pointgnn} for both cars and pedestrians; CasA \cite{casa} for cars; and VirConv \cite{virconv} for cars, respectively.
Note that CasTrack operates in an online tracking mode to ensure a fair comparison, while the results of other competitors are obtained from the KITTI benchmark.
The ``C'' and ``L'' represent the camera and LiDAR sensors, respectively.
The best results are in bold. 
}
\label{tab1}
\centering
\renewcommand{\arraystretch}{1.5}
\scalebox{0.97}{
\begin{tabular}{c|c|cccccc|cccccc}
\hline
\multirow{2}{*}{Method} & \multirow{2}{*}{Input} & \multicolumn{6}{c|}{Car} & \multicolumn{6}{c}{Pedestrian} \\ \cline{3-14} & & HOTA & DetA & AssA & MOTA & IDWS & sMOTA & HOTA & DetA & AssA & MOTA & IDWS & sMOTA \\ \hline
FAMNet \cite{FAMNet}                       & C     & 52.56 & 61.00 & 45.51 & 75.92 & 521  &59.02   & -     &  -    &  -    &   -   &  -  &   -       \\
LGM \cite{lgm}                             & C     & 73.14 & 74.61 & 72.31 & 87.60 & 448  &72.76   & -     &  -    &  -    &   -   &  -  &   -       \\
DEFT \cite{lgm}                            & C     & 74.23 & 75.33 & 73.79 & 88.38 & 344  &74.04   & -     &  -    &  -    &   -   &  -  &  -         \\
TripletTrack\cite{triplettrack}            & C     & 73.58 & 73.18 & 74.66 & 84.32 & 522  &72.26   & 42.77 & 39.54 & 46.54 & 50.08 & 323 & \textbf{35.71}  \\  
PolarMOT\cite{polarmot}                    & L     & 75.16 & 73.94 & 76.95 & 85.08 & 462  &71.82   & 43.59 & 39.88 & \textbf{48.12} & 46.98 & 270 & 24.61     \\
FANTrack\cite{fantrack}                    & L     & 60.85 & 64.36 & 58.69 & 75.84 & 743  &61.49   & -     & - &   -  &   -  &  -   &    -       \\
BeyondPixels\cite{beyondpixels}            & C+L   & 63.75 & 72.87 & 56.40 & 82.68 & 934  &69.98   & -     & -  &   -  &   -  &  -   &   -     \\
mmMOT\cite{mmmot}                          & C+L   & 62.05 & 72.29 & 54.02 & 83.23 & 733  &70.12   & -     &    -  &   -  &   -  &   -  &     -      \\
AB3DMOT\cite{ab3dmot}                      & C+L   & 69.99 & 71.13 & 69.33 & 83.61 & 113  &70.80   & 37.81 &32.37    & 44.33   & 38.13    &   \textbf{181}   &   20.80         \\
JRMOT\cite{jrmot}                          & C+L   & 69.61 & 73.05 & 66.89 & 85.10 & 271  &72.11   & 34.24 &38.79    &  30.55    &  45.31    &  631    &   29.78        \\
MOTSFusion \cite{motsfusion}               & C+L   & 68.74 & 72.19 & 66.16 & 84.24 & 415  &71.14   & -     &   -   &  -    &  -    &  -    &   -         \\
JMODT\cite{jmodt}                          & C+L   & 70.73 & 73.45 & 68.76 & 85.35 & 350  &72.19   & -     &  -    &  -    & -     &  -    & -          \\
DeepFusionMOT\cite{deepfusionmot}          & C+L   & 75.46 & 71.54 & 80.05 & 84.63 & 84   &71.66   & -     &  -    &  -    & -     &  -    &   -         \\
YONTD-MOT \cite{YONTD-MOT}                 & C+L   & 78.08 & 74.16 & 82.84 & 85.09 & \textbf{42} & 73.55  & 25.89 & 27.31 & 25.02 & 26.19 & 1068 & 11.34  \\
StrongFusionMOT \cite{sfmot}              & C+L   & 75.65 & 72.08 & 79.84 & 85.53 & 58   & 72.62  & 43.42 & 38.86 &  48.83    & 39.04     &  316    & 16.54          \\
BcMODT \cite{BcMODT}                       & C+L   & 71.00 & 73.62 & 69.14 & 85.48 & 381  & 72.22  & -     &  -    &  -    & -     &  -    & -      \\ \hdashline
EagerMOT$^{C1L1}$ \cite{eagermot}     & C+L   & 74.39 & 75.27 & 74.16 & 87.82 & 239  &74.97   & 39.38 & 40.60     &  38.72    & 49.82     & 496     &  28.01          \\ 
Ours$^{C1L1}$                                 & C+L   & 78.43 & 76.02 & 81.56 & 89.72 & 82  &76.27   & \textbf{44.73} & \textbf{44.00} & 45.90 & \textbf{56.63} & 294 & 33.21    \\
$\Delta$                                      & -     & \textbf{+4.04} & \textbf{+0.75} & \textbf{+7.40} & \textbf{+1.90} & \textbf{+157} & \textbf{+1.30} &\textbf{+5.35} &\textbf{+3.40} &\textbf{+7.18} &\textbf{+6.81} &\textbf{+202} &\textbf{+5.20}  \\  \hdashline
CasTrack$^{L2}$ \cite{CasTrack}    & L     & 77.32 & 75.08 & 80.32 & 86.33 & 184  & 74.27   & -   &  -    &  -    & -     &  -    & -      \\ 
Ours$^{C2L2}$                                 & C+L   & 80.87 & 78.77 & 83.66 & 91.48 & 71 & 78.97 & -   &  -    &  -    & -     &  -    & -      \\ 
$\Delta$                                      & -     & \textbf{+3.55} & \textbf{+3.69} & \textbf{+3.34} & \textbf{+5.15} & \textbf{+113}   & \textbf{+4.70} & -   &  -    &  -    & -     &  -    & -    \\  \hdashline
CasTrack$^{L3}$ \cite{CasTrack}    & L     & 79.97 & 77.94 & 82.67 & 89.19 & 201 & 77.07   & -   &  -    &  -    & -     &  -    & -       \\ 
Ours$^{C3L3}$                               & C+L   & \textbf{82.04}& \textbf{79.86}& \textbf{84.90} & \textbf{91.77} & 63 & \textbf{79.86} & -   &  -    &  -    & -     &  -    & -       \\ 
$\Delta$                                      & -     & \textbf{+2.07} & \textbf{+1.92} & \textbf{+2.23} & \textbf{+2.58} & \textbf{+138}   & \textbf{+2.79} & -   &  -    &  -    & -     &  -    & -    \\
\hline
\end{tabular}
}
\end{table*}

\section{Experiment}
\subsection{Dataset} 
To evaluate the performance of our CrossTracker, we utilize the widely recognized KITTI tracking benchmarks.
The KITTI dataset comprises 21 training sequences and 29 test sequences, totaling 8008 and 11095 frames, respectively. Each frame includes a point cloud and an RGB image.
We evaluate our CrossTracker on the \textit{Car} and \textit{Pedestrian} categories.

\subsection{Camera- and LiDAR-based Object Detectors}
To ensure a fair and comprehensive evaluation, we experiment with various combinations of Camera- and LiDAR-based detectors (see Table \ref{tab4}).
Specifically, we incorporate LiDAR-based detectors such as Point-GNN \cite{pointgnn}, PointRCNN \cite{pointrcnn}, CasA \cite{casa} and VirConv \cite{virconv}, and utilize RRC \cite{rrc}, Perma \cite{PermaTrack}, and Track-RCNN \cite{trackrcnn} as camera-based detectors.

\subsection{Evaluation Metrics} \label{Evaluation Metrics}
The official KITTI evaluation toolkit \cite{kitti} is employed to evaluate the performance of CrossTracker quantitatively. CLEAR MOT \cite{clear_mot} is a standard evaluation rule used for the MOT task, such as Multi-Object Tracking Accuracy (MOTA), scaled MOTA (sMOTA), Multi-Object Tracking Precision (MOTP), and ID Switch (IDSW). Apart from the CLEAR metrics, HOTA \cite{hota}, a recently proposed MOT evaluation metric, is currently used by the KITTI dataset as one of the main evaluation metrics for tracking performance. As an evaluation metric that unifies detection and association quality, HOTA can be decomposed into several sub-metrics, mainly including Detection Accuracy (DetA) and Association Accuracy (AssA). Thus, the CLEAR and HOTA metrics are employed for MOT performance evaluation. 

\subsection{Experimental Setup}
CrossTracker is developed using Python and PyTorch, and the experiments are carried out on a desktop computer equipped with an Intel Core i9 3.70GHz CPU, 128GB of RAM, and a GTX 4090 GPU. 
Besides, we split the training set of KITTI into two parts: the training set containing sequences \textit{0000, 0002, 0003, 0004, 0005, 0007, 0009, 0011, 0017, 0020} and the validation set containing sequences \textit{0001, 0006, 0008, 0010, 0012, 0013, 0014, 0015, 0016, 0018, 0019}. This enables us to train the M$^{3}$ module and validate CrossTracker.
During training, the epoch is 10, the learning rate is 0.0001, and the weight decay is 0.00005.

\begin{figure*}[t]
 \centering
  \includegraphics[width=\textwidth]{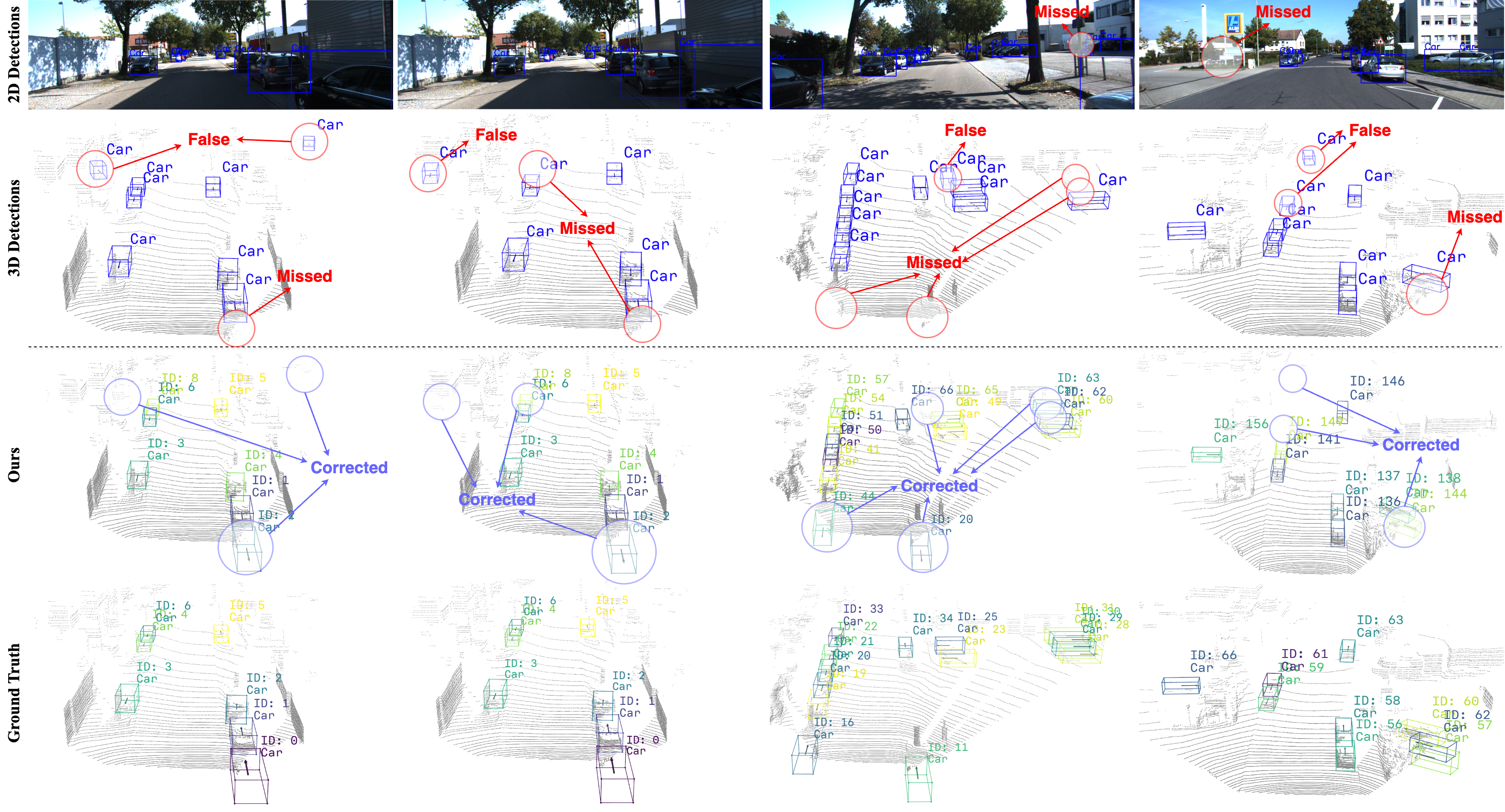}
  \caption{\textbf{Qualitative evaluation of our CrossTracker on KITTI.} 
  It presents a qualitative comparison of the sequence \textit{0001} of the KITTI validation dataset.
  The first row displays detections from the camera-based detector RRC\cite{rrc}, the second row shows detections from the LiDAR-based detector Point-GNN\cite{pointgnn}, the third row demonstrates tracking results from our CrossTracker, and the fourth row provides the ground truth.}
  \label{fig:results vis}
\end{figure*}

\subsection{Main Results}
\subsubsection{Quantitative Evaluation} 
We employ two baseline methods, EagerMOT \cite{eagermot} and CasTrack \cite{CasTrack}, and conduct comparative experiments on the KITTI test set using three different combinations of camera- and LiDAR-based detectors to highlight the superiority of our CrossTracker.

As shown in Table \ref{tab1}, CrossTracker consistently outperforms EagerMOT\cite{eagermot} across all metrics, particularly for the challenging \textit{Pedestrian} category.
Specifically, for the \textit{Car} category, CrossTracker improves HOTA, DetA, AssA, MOTA, IDWS, and sMOTA by 4.04\%, 0.75\%, 7.40\%, 1.90\%, 157, and 1.30\% respectively.
Similarly, for the \textit{Pedestrian} category, CrossTracker improves HOTA, Det, AssA, MOTA, IDWS, and sMOTA by 5.35\%, 3.40\%, 7.18\%, 6.81\%, 202, and 5.20\%, respectively. 
Importantly, both CrossTracker and EagerMOT utilize the same 2D detections from RRC\cite{rrc} and Track-RCNN\cite{trackrcnn} and the same 3D detections from Point-GNN\cite{pointgnn} as inputs. However, EagerMOT incorporates additional 2D segmentation information, while CrossTracker does not.
Despite this, CrossTracker remains significantly superior to EagerMOT. This demonstrates the effectiveness of our two-stage framework in tackling the four challenges shown in Fig. \ref{fig:problems}, particularly (c) and (d) which EagerMOT cannot handle.

Additionally, we compare CrossTracker with the state-of-the-art CasTrack \cite{CasTrack}, demonstrating that the cross-correction in our TR module significantly improves 3D trajectory output.
Specifically, both CrossTracker and CasTrack utilize the same 3D detections as input, sourced from either CasA \cite{casa} or VirConv \cite{virconv}. However, CrossTracker additionally leverages 2D detections from Perma \cite{PermaTrack} as an auxiliary data source.

Our experimental results demonstrate that our method surpasses CasTrack in online performance. Notably, when using either Casa or VirConv as LiDAR detectors, our method achieves a 3.55\% and 2.07\% improvement in the HOTA metric, respectively. Furthermore, our method remains highly competitive compared to sixteen other state-of-the-art methods, particularly in terms of the HOTA metric.

\subsubsection{Qualitative Evaluation} 
In addition to the quantitative results on the KITTI dataset, Fig. \ref{fig:results vis} provides a visual representation of CrossTracker's capability in correcting a significant number of false and missed detections in both camera and LiDAR data streams.
For example, in the four columns shown, false LiDAR detections are corrected by the camera stream, while missed LiDAR detections in the first and second columns are also recovered with camera information. Conversely, the fourth column demonstrates how camera-missed detections are corrected using LiDAR data.
Even in the challenging case of the third column, where both LiDAR and camera streams miss the car on the far right, CrossTracker is able to accurately track it.
These results underscore CrossTracker's exceptional ability to overcome complex detection challenges, demonstrating its robustness and effectiveness in real-world scenarios.

\begin{table}[h]
\caption{A ablation study on the performance of the classifier in the M$^3$ module for estimating the consistency between two objects under different modality features on the KITTI \textit{validation} set.}
\label{tab2-1}
\centering
\renewcommand{\arraystretch}{1.5}
\scalebox{1}{
\begin{tabular}{ccc|ccc}
\hline
IFM & GFM & PFM & F1\_score    & Precision & Recall \\ \hline
\checkmark    &             &            & 94.58          & 94.69          & 94.58          \\
\checkmark    & \checkmark  &            & 96.85          & 97.00          & 96.85          \\
\checkmark    & \checkmark  & \checkmark & \textbf{96.90} & \textbf{97.07} & \textbf{96.90} \\ \hline
\end{tabular}
}
\end{table}

\begin{table*}[h]
\caption{
Ablation study of modal features impact on the M$^3$ module on the KITTI \textit{validation} set.
Experiments use the RRC \cite{rrc} and Track-RCNN \cite{trackrcnn} as camera-based detectors and Point-GNN \cite{pointgnn} as a LiDAR-based detector.
}
\label{tab2}
\centering
\renewcommand{\arraystretch}{1.5}
\scalebox{1}{
\begin{tabular}{ccc|ccc|ccc}
\hline
\multirow{2}{*}{IFM} & \multirow{2}{*}{GFM} & \multirow{2}{*}{PFM}  & \multicolumn{3}{c|}{Car}   & \multicolumn{3}{c}{Pedestrian}     \\ \cline{4-9}
                     &                      &                       & HOTA           & DetA           & AssA           & HOTA           & DetA           & AssA           \\ \hline
\checkmark           &                      &                       & 82.76          & 82.42          & 83.38          & 50.28          & 50.29 & 50.60          \\
\checkmark           & \checkmark           &                       & 83.07          & 82.48          & 83.93          & 50.14          & 49.97          & 50.63   \\
\checkmark           & \checkmark           & \checkmark           & \textbf{83.78} & \textbf{82.54} & \textbf{85.31} & \textbf{51.91} & \textbf{50.55}          & \textbf{53.63} \\ \hline
\end{tabular}
}
\end{table*}

\subsection{Ablation Study}
To systematically analyze the contributions of the proposed M$^3$ and TR modules, and the influence of various object detectors on the performance of CrossTracker, we conduct ablation experiments on the KITTI validation set, adhering to the official KITTI evaluation tool.

\subsubsection{Effects of M$^3$} 
The results in Table \ref{tab2-1} demonstrate that each modality (IFM, GFM, and PFM) significantly contributes to estimating the inter-object consistency probability.
Specifically, when only image modality is available, \textit{Classifier} achieves the lowest performance with \textit{F1\_score}, \textit{Precision}, and \textit{Recall} of 94.58\%, 94.69\%, and 94.58\%, respectively.
Incorporating plane geometric features improves these metrics by 2.27\%, 2.31\%, and 2.27\%, respectively. The addition of point cloud features yields the highest performance, with the metrics reaching 96.90\%, 97.07\%, and 96.90\%, respectively.

Table \ref{tab2} further demonstrates that all three modalities significantly improve the performance of 3D MOT. Although Table \ref{tab2-1} shows that point cloud features have limited improvement on the performance of the \textit{Classifier}, they exhibit significant performance gains in the 3D MOT task. 
Specifically, when using only the image modality, the HOTA, DetA, and AssA metrics for the \textit{Car} category are 82.76\%, 82.42\%, and 83.38\%, respectively, while for the \textit{Pedestrian} category, these metrics are 50.28\%, 50.28\%, and 50.60\%. 
Furthermore, after adding plane geometry features, the HOTA, DetA, and AssA metrics for the \textit{Car} category are further improved by 0.31\%, 0.06\%, and 0.55\%, respectively. However, for the \textit{Pedestrian} category, the impact of adding plane geometry features on 3D MOT performance is minimal. 
Finally, after adding point cloud features, the performance of both the \textit{Car} and \textit{Pedestrian} categories has been significantly improved. Specifically, on the \textit{Car} category, the HOTA, DetA, and AssA metrics are improved by 0.71\%, 0.06\%, and 1.38\%, respectively, while on the \textit{Pedestrian} category, the HOTA, DetA, and AssA metrics are improved by 1.77\%, 0.58\%, and 3.00\%.

\begin{table*}[ht]
\caption{
Ablation study results on the KITTI validation set for the three steps within the TR module addressing cases (a), (b), (c), (d), and (e) in Fig. \ref{fig:examples of cr}.
Camera-based detectors, RRC \cite{rrc} for cars and Track-RCNN \cite{trackrcnn} for pedestrians, and a LiDAR-based detector, Point-GNN \cite{pointgnn} for both cars and pedestrians, are employed in the experiments.
}
\label{tab3}
\centering
\renewcommand{\arraystretch}{1.5}
\scalebox{1}{
\begin{tabular}{c|ccccc|ccc|ccc}
\hline
\multirow{2}{*}{Usage of Streams} & \multicolumn{5}{c|}{Cross Correction}                     & \multicolumn{3}{c|}{Car}                         & \multicolumn{3}{c}{Pedestrian}                   \\ \cline{2-12} 
                                  & (a)       & (b)       & (c)       & (d)       & (e)       & HOTA           & DetA           & AssA           & HOTA           & DetA           & AssA           \\ \hline
LiDAR Stream                      &           &           &           &           &           & 78.08          & 78.11          & 78.31          & 48.93          & 47.54          & 50.69          \\ \hdashline
\multirow{5}{*}{Dual Stream}      & \checkmark &           &           &           &           & 79.39          & 80.34          & 78.71          & 50.16          & 50.36          & 50.30          \\
                                  & \checkmark & \checkmark &           &           &           & 79.64          & 80.62          & 78.93          & 50.18          & 50.38          & 50.32          \\
                                  & \checkmark & \checkmark & \checkmark &           &           & 81.76          & 82.15          & 81.62          & 51.90          & \textbf{50.61}          & 53.55          \\
                                  & \checkmark & \checkmark & \checkmark & \checkmark &           & 81.93          & 82.33          & 81.77          & \textbf{51.92} & 50.60 & 53.59          \\
                                  & \checkmark & \checkmark & \checkmark & \checkmark & \checkmark & \textbf{83.78} & \textbf{82.54} & \textbf{85.31} & 51.91          & 50.55          & \textbf{53.63} \\ \hline
\end{tabular}
}
\end{table*}

\begin{table*}[ht]
\caption{Ablation study of different camera- and LiDAR-based detectors on the $Car$ category of the KITTI \textit{validation} set.}
\label{tab4}
\centering
\renewcommand{\arraystretch}{1.5}
\scalebox{1}{
\begin{tabular}{cc|ccccccccc}
\hline
LiDAR-based detector & Camera-based detector & HOTA  & DetA   & AssA  & MOTA  & sMOTA  & MT     & ML   & FRAG  & IDSW  \\ \hline
PointRCNN \cite{pointrcnn}    & RRC \cite{rrc}        & 80.65 & 79.64 & 81.96 & 92.91 & 79.75 & 171    & 2 & 38  & 15   \\
Point-GNN \cite{pointgnn}     & RRC \cite{rrc}        & 83.78 & 82.54 & 85.31 & \textbf{94.71} & 82.79 & 171 & 2 & 33  & 9      \\ 
CasA \cite{casa}              & RRC \cite{rrc}        & 85.34 & 84.00 & 86.87 & 93.32 & 83.70 & \textbf{170} & \textbf{1} & 31  & 12     \\
VirConv \cite{virconv}        & RRC \cite{rrc}        & \textbf{88.21} & \textbf{86.93} & \textbf{89.62} & 94.12 & \textbf{86.30} & 171 & \textbf{1} & \textbf{19}  & \textbf{6}    \\ \hdashline
PointRCNN \cite{pointrcnn}    & Perma \cite{PermaTrack}        & 80.03 & 78.35 & 82.02 & 92.12 & 77.83 & \textbf{172} & \textbf{0} & 29  & 14   \\
Point-GNN \cite{pointgnn}     & Perma \cite{PermaTrack}        & 83.96 & 82.90 & 85.29 & \textbf{95.23} & 83.08 & 174 & \textbf{0} & 27  & 10    \\ 
CasA \cite{casa}              & Perma \cite{PermaTrack}        & 85.38 & 83.93 & 87.02 & 93.20 & 83.49 & 173 & \textbf{0} & 21  & \textbf{5}     \\
VirConv \cite{virconv}        & Perma \cite{PermaTrack}        & \textbf{87.89} & \textbf{86.56} & \textbf{89.35} & 93.63 & \textbf{85.78}  & 174 & \textbf{0} & \textbf{15}  & \textbf{5}  \\ \hline
\end{tabular}}
\end{table*}

\subsubsection{Effects of TR}
The results presented in Table \ref{tab3} clearly demonstrate that the two-stage tracking paradigm with cross correction outperforms the single-stage paradigm. Moreover, the sequential addressing of the five challenging cases illustrated in Fig. \ref{fig:examples of cr} by the TR module significantly contributes to the enhancement of the final 3D trajectory output.
Specifically, when using only the LiDAR stream for single-stage 3D MOT, the tracking performance for both the \textit{Car} and \textit{Pedestrian} categories is the lowest, with HOTA metrics of 78.08\% and 48.93\%, respectively.
Subsequently, addressing cases (a) to (e) sequentially leads to significant improvements in all tracking metrics for the textit{Car} category. Notably, the HOTA metric increases by 1.31\%, 0.25\%, 2.12\%, 0.17\%, and 1.85\%, respectively. This highlights the substantial impact of addressing cases (a), (c), and (e) on the overall performance.
In addition, while the \textit{Pedestrian} category also benefits from the proposed TR module, the most significant enhancements are observed after addressing cases (a) and (c), resulting in HOTA metric improvements of 1.23\% and 1.72\%, respectively. It's worth noting that the handling of case (e) leads to a slight decrease of 0.01\% and 0.05\% in HOTA and DetA, respectively, for the \textit{Pedestrian} category. This minor decrease might be attributed to the misidentification of simultaneous terminations as missed detections in both modalities.
However, building upon the handling of case (d), further addressing case (e) yields a significant improvement of 1.85\% in HOTA for the textit{Car} category, making the inclusion of case (e) in the TR module worthwhile.

\subsubsection{Effects of Camera- and LiDAR-based Object Detectors}
Based on the tracking-by-detection paradigm, CrossTracker can be used in conjunction with arbitrary camera- and LiDAR-based detectors. 
CrossTracker, rooted in the tracking-by-detection paradigm, is compatible with various camera- and LiDAR-based detectors.
To comprehensively analyze the effect of different camera- and LiDAR-based detectors on CrossTracker, we conduct eight sets of comparative experiments (see Table~\ref{tab4}) using PointRCNN \cite{pointrcnn}, Point-GNN \cite{pointgnn}, CasA \cite{casa} and VirConv \cite{virconv} as LiDAR-based detectors, and employ RCC \cite{rrc} and Perma \cite{PermaTrack} as camera-based detectors. 
Note that for the six detectors, the detection results provided by the corresponding authors are directly employed.

Table~\ref{tab4} presents the results of our experiments demonstrating that while the choice of camera- and LiDAR-based detectors can influence the overall performance of CrossTracker, our proposed two-stage paradigm consistently yields more robust and smoother 3D trajectories as indicated by lower IDWS scores.
Our findings also highlight the significant impact of the quality of input 3D detection results on the final 3D tracking outcomes. Specifically, VirConv, when paired with either RRC or Perma, consistently outperformed other detector combinations, achieving the highest HOTA metrics of 88.21\% and 87.89\%, respectively. CasA followed closely, yielding HOTA metrics of 85.34\% and 85.38\% when combined with RRC and Perma, respectively.

\begin{figure*}[t]
 \centering
  \includegraphics[width=\textwidth]{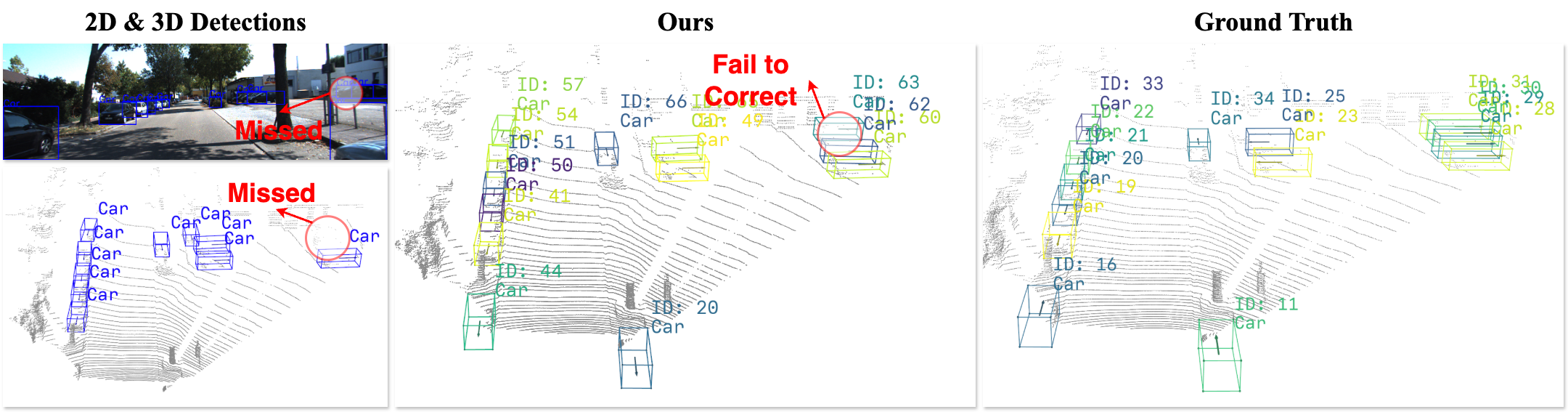}
  \caption{\textbf{The case of failure.} 
  This is a failure case where two modalities simultaneously miss detection at the boundary of the field of view, and the failure has not been successfully corrected by our CrossTracker.}
  \label{fig:case of failure}
\end{figure*}

\section{Limilation}
Our CrossTracker, while effective, has some limitations.
Firstly, tracking objects missed by both camera and LiDAR streams at the field of view boundary is difficult. As illustrated in Fig.~\ref{fig:case of failure}, both modalities failed to detect a car located at the view boundary, leading to tracking failure.
Secondly, the TR module may introduce noise during the cross correction of camera and LiDAR trajectories, which can adversely affect the output 3D trajectories. As shown in Table \ref{tab3}, the handling of case (e) slightly decreases the HOA and DetA metrics for the \textit{Pedestrian} category.
Thirdly, our method is limited to single-camera (monocular) input and does not yet support multi-camera setups.
In light of these limitations, we are working to enhance the robustness and generalizability of CrossTracker.

\section{Conclusion}
We present a novel two-stage multi-modal 3D MOT method, dubbed CrossTracker, inspired by the coarse-to-fine paradigm commonly employed in 3D object detection.
By leveraging a two-stage tracking paradigm, CrossTracker effectively addresses common tracking failures often caused by detection problems in complex 3D MOT scenarios.
A key component of CrossTracker is the M$^3$ module, which integrates the modeling of multimodal features (including images, point clouds, and even image-derived plane geometry) and the estimation of consistency probabilities among objects into a unified end-to-end network.
This module eliminates the need for manual calculations and significantly advances the state-of-the-art.
Building upon the M$^3$ module, the C-TG module generates initial coarse dual-stream trajectories. Subsequently, the TR module refines these trajectories through cross correction between camera and LiDAR modalities.
This dual-stream refinement process leads to highly robust 3D trajectories that surpass the performance of existing methods.

Our work highlights the effectiveness of a two-stage, coarse-to-fine paradigm for 3D MOT, opening up new avenues for robust 3D MOT solutions. Moreover, our proposed CrossTracker framework not only sets a new benchmark in this field but also serves as a source of inspiration for future research. We believe this work will spark a wave of follow-up studies and drive the advancement of 3D MOT research.

\bibliographystyle{IEEEtran}
\bibliography{acmart}

\newpage

\vfill

\end{document}